\documentclass[reprint,amsmath,amssymb,aps]{revtex4-2}

\usepackage{graphicx}
\usepackage{dcolumn}
\usepackage{bm}
\usepackage{hyperref}
\usepackage[mathlines]{lineno}
\usepackage{color}
\usepackage{amsmath,amsfonts,amssymb}
\usepackage{float}
\usepackage{verbatim}
\usepackage{bm}
\usepackage{mathtools}

\usepackage[title]{appendix}

\newcommand*\mat[1]{\mathbf{#1}}
\newcommand{\J}[2]{\mathbf{#1}_{#2}}
\newcommand{\MB}[1]{\mat{\bar{#1}}}
\newcommand{\MT}[1]{\mat{\tilde{#1}}}

\usepackage{color}
\definecolor{darkblue}{rgb}{0.,0.1,0.6}
\definecolor{darkcyan}{rgb}{0.1,0.3,0.4}
\definecolor{darkgreen}{rgb}{0,0.4,0}
\definecolor{darkred}{rgb}{0.6,0,0}
\definecolor{lightgrey}{rgb}{0.6,0.7,0.8}

\begin{document}
\title{Recurrent Neural Networks for Partially Observed Dynamical Systems}
\author{Uttam Bhat}
\affiliation{University of California, Santa Cruz, CA, 95064, United States}
\email{ubhat@ucsc.edu}
\author{Stephan B. Munch}
\affiliation{Applied Mathematics, University of California, Santa Cruz, CA, 95064, United States}
\email{smunch@ucsc.edu}

\begin{abstract}
    Complex nonlinear dynamics are ubiquitous in many fields. Moreover, we rarely have access to all of the relevant state variables governing the dynamics.  Delay embedding allows us, in principle, to account for unobserved state variables.  Here we provide an algebraic approach to delay embedding that permits explicit approximation of error. We also provide the asymptotic dependence of the first order approximation error on the system size. More importantly, this formulation of delay embedding can be directly implemented using a Recurrent Neural Network (RNN).  This observation expands the interpretability of both delay embedding and RNN and facilitates principled incorporation of structure and other constraints into these approaches.
\end{abstract}
\date{\today}

\maketitle

\section{Introduction}
\label{intro}

Forecasting dynamical systems is important in many disciplines. Weather and climate \cite{palmer2001nonlinear}, ecology \cite{lek1996application,luo2011ecological}, biology, \cite{lachowicz2011individually,imoto2003bayesian}, fluid dynamics \cite{farmer1987predicting} etc. are generally modeled with nonlinear, discrete time equations or continuous time differential equations. In many cases, these non-linear systems are chaotic and subject to stochastic drivers. However, the empirical data available are often incomplete; It is common to observe only a subset of the state variables or measure some coarse-grained statistic of the underlying state. In such a situation, all hope is not lost; Takens embedding theorem \cite{takens1981detecting} shows that time-delayed versions of a single observable can be used in place of the unobserved dimensions to reconstruct the attractor manifold permitting accurate short- and mid- term forecasts \cite{sugihara1990nonlinear}. 

Takens theorem shows that any universal function approximator (given enough data) would be able to infer the function mapping the delay state vector to its future value. However, the proof of Takens' theorem is topological and non-constructive. Therefore, one approach to reconstruct dynamics using partial state variables is accomplished by using off-the-shelf function approximation methods to infer the function mapping the delay vector to its future values from time series data.

Due to the recent developments in machine learning, there are abundant choices for tools to perform nonlinear regression. The common candidates are local linear regressions, neural networks and Gaussian processes \cite{ragwitz2002markov,garcia2005multivariate,qin2019data,raissi2018multistep,lusch2018deep}. Recurrent Neural Networks (RNN) and its variants are some of the most widely used tools for time series prediction. Although RNNs have been successfully applied to forecasting in a wide range of problems, literature on the mathematical reasons why they work so well is largely lacking \cite{weinan2020towards,caterini2018deep}.
Early justifications for using a recurrent architecture include 1) being able to store temporal information \cite{mandic2001recurrent}, 2) neural networks with feedback capture time dependencies better, 3) are natural candidates for nonlinear autoregressive models \cite{connor1994recurrent}, 4) leads to reduced number of parameters due to weight sharing \cite{goodfellow2016deep}. RNNs were also used for time series prediction due to their ability to be continually trained \cite{williams1989learning}

Neural network architectures for nonlinear dynamics are generally benchmarked on large data applications. Although there are asymptotic results proving the efficacy of some of these architectures, these results are not useful in many real world use cases where data can be limiting. Neural networks are also known to require a high-degree of application-specific hyperparameter tuning \cite{goodfellow2016deep}. This makes it hard to use neural networks where there isn't sufficient data for a dedicated validation dataset. Progress can be made by assuming smoothness of the underlying functions to obtain less stringent requirements on the size of data needed to embed high-dimensional dynamics \cite{cheng1994orthogonal,munch2017circumventing}.

Here we present a new approach to delay embedding through simple algebraic manipulation of the dynamical equations. We derive an approximation to the delay dynamics in terms of the original dynamics. We hypothesize that this approximation allows us to infer the delay map more efficiently with less data due to two reasons - 1) the original dynamics will be smoother than the delay function due to the distortions introduced by folding of the attractor manifold \cite{kantz1997scalar}, and 2) the original dynamics are generally of smaller dimensionality than the delay function. We also discuss how our approximation is amenable to mechanistic interpretation unlike traditional delay-embedding and nonlinear autoregressive models. 

We use this approximation to encode a Recurrent Neural Network (RNN) to accomplish forecasting chaotic dynamics. Connections between dynamical system and RNN have been made in the past \cite{elman1990finding,han2004modeling,zimmermann2000modeling}. However, these connections do not take advantage of the recurrent nature of partially observed dynamics. 

In general data-driven function approximation methods work well when the time series data has a wide coverage across the domains of the functions. This is truly achieved when the dynamics are ergodic. In a practical setting, chaos or stochasticity too can be sufficient to achieve this. 

In the next section, we calculate the first order error due to partial observation of a system. We then develop a recursive approximation of the dynamics using only the observed states, and calculate the first-order expansion of the covariance of the recursion error.
In section \ref{rnn}, we develop a recurrent neural network architecture that uses the recursive structure of the dynamics of the observed states. In section \ref{illustrative_example}, we illustrate the effects of partial observation on the delay dynamics using both an analytically solvable example and more complex dynamics commonly used in biophysical systems.
We use short simulated time-series as the effectiveness of the RNN structure over feed-forward networks is most evident when the number of training points is less than a hundred.
Finally, we discuss the potential for using other function approximators to take advantage of the general structure of dynamics to achieve more efficient representations of data.

\section{Recursive expansion of dynamics}
\label{error}

Assume the dynamics are completely represented with a system of $M$ state variables, say $\mat{z_t}=\left\{z_{1,t},  z_{2,t},...,z_{M,t}\right\}^T$ and the dynamics are given by 
\begin{align} \label{starting_point}
  \frac{dz_{1}}{dt}&=f_{1}\left(\mat{z}_{t}\right) \\
        & \vdots\notag \\
 \frac{dz_{M}}{dt}&=f_{M}\left(\mat{z}_{t}\right) \notag
\end{align}

\noindent However, the subsequent arguments are more transparent in discrete time, so we work with the corresponding flow map integrated on a unit time step, $z_{1,t}=F_1\left(\mat{z}_{t-1} \right)$,... $z_{M,t}=F_M\left(\mat{z}_{t-1} \right)$ which we write compactly as $\mat{z}_{t}=\mat{F}\left(\mat{z}_{t-1} \right)$  

Since our focus is on partially observed systems, we split the state variables $\mathbf{z}_t$ into two subsets:  $\mat{x}_{t}=\left\{ z_{1,t},\,z_{2,t},\,\ldots,\,z_{n,t}\right\}^T$ representing the observed state variables and $\mat{y}_{t}=\left\{ z_{n+1,t},\,\ldots,\,z_{M,t}\right\}^T$ containing the remaining, unobserved state variables.  We re-write the dynamics as 

\begin{align} \label{two_states}
\mat{x}_{t}= \mat{F}\left(\mat{x}_{t-1},\mat{y}_{t-1}\right)\nonumber\\
\mat{y}_{t}= \mat{G}\left(\mat{x}_{t-1},\mat{y}_{t-1}\right)
\end{align}

\noindent where $\mat{F}$ represents the maps for the $n$ observed states and $\mat{G}$ represents the maps for the $M-n$ unobserved states. 

There are several ways to proceed, including (i) implicitly accounting for the unobserved states using time lags (\cite{Deyle2013,Munch2018}), or (ii) modeling the complete dynamics and imputing the unobserved states using a hidden Markov approach (e.g. \cite{Morales2004a}).  However, (ii) requires that we have a reasonable model for the complete state dynamics and significant problems arise when the model is inaccurate. Since we assume the complete dynamics are unknown, we focus on (i).  

As a first step to doing this, we shift the map for the unobserved states back by one time step and substitute this into the dynamics for the observed states.  

\begin{align}
\mat{x}_{t}&=\mat{F}\left(\mat{x}_{t-1},\mat{y}_{t-1}\right) \nonumber\\
&=\mat{F}\left(\mat{x}_{t-1},\mat{G}\left[\mat{x}_{t-2},\mat{y}_{t-2}\right]\right) \nonumber\\
&=E_{\mat{y}_{t-2}}\left[ \mat{F}\left(\mat{x}_{t-1},\mat{G}\left[\mat{x}_{t-2},\mat{y}_{t-2}\right]\right) \vert \mat{x}_{t-1},\mat{x}_{t-2}\right] + \varepsilon_t \nonumber\\
&\approx\mat{F}\left(\mat{x}_{t-1},\mat{G}\left[\mat{x}_{t-2},\MB{y}^{t-2}\right]\right)+\varepsilon_t \nonumber\\
\label{apx2}
\end{align}

\noindent where $\MB{y}^{t-2}=E\left[\mat{y}|\mat{x}_{t-1},\mat{x}_{t-2}\right]$ is the conditional expectation for $\mat{y}$ given the current and previous observation for $\mat{x}$.  The apparent process noise $\varepsilon_t$ is given by $\varepsilon_t=\mat{F}\left[\mat{x}_{t-1},\mat{G}\left(\mat{x}_{t-2},\mat{y}_{t-2}\right)\right]-\mat{F}\left[\mat{x}_{t-1},\mat{G}\left(\mat{x}_{t-2},\MB{y}^{t-2}\right)\right]$. The approximation in line 4 of eq.~\eqref{apx2} assumes $\mat{F}$ and $\mat{G}$ are almost linear for simplicity.

We can continue along this path an arbitrary number of times, each iteration adding another lag of $\mat{x}$ and pushing back the dependence on $\mat{y}$. Doing so $d$ times we get,

\begin{align}
\mat{x}_{t}&=\mat{F}\left( \mat{x}_{t-1},\mat{G}\left[ \mat{x}_{t-2},\hdots \mat{G}\left\{\mat{x}_{t-d},\MB{y}^{t-d}\right\} \hdots \right] \right)+\varepsilon_t \label{apxd1}\\
&=\MT{F}_d\left(\mat{x}_{t-1},\hdots, \mat{x}_{t-d}\right)+\mat{\varepsilon}_t \label{apxd2}
\end{align}

\noindent where, in keeping with the previous notation, $\MB{y}^{t-d}=E\left[\mat{y}_{t-d}|\mat{x}_{t-1},\hdots,\mat{x}_{t-d}\right]$ and $\varepsilon_t=\mat{F}\left( \mat{x}_{t-1},\mat{G}\left[ \mat{x}_{t-2},\hdots \mat{G}\left\{\mat{x}_{t-d},\mat{y}_{t-d}\right\} \hdots \right] \right)-\mat{F}\left( \mat{x}_{t-1},\mat{G}\left[ \mat{x}_{t-2},\hdots \mat{G}\left\{\mat{x}_{t-d},\MB{y}^{t-d}\right\} \hdots \right] \right)$. As we show in the illustrative example in section \ref{illustrative_example}, we expect the dependence of function $\hat{\mat{F}}$ on $\mat{F}$ and $\mat{G}$ to be complicated. Therefore it is hard to connect the function $\hat{\mat{F}}$ with the parameters of the generators of the dynamics, $\mat{F}$ and $\mat{G}$. With our recursive approximation \eqref{apxd1}, we can retain the identity of the ground-truth dynamics $\mat{F}$ and $\mat{G}$ using our approximation \eqref{apxd1}. 

To provide a benchmark for our approximation, we estimate $\varepsilon$ in the limit of large data. We simulate the exact dynamics for 30000 time steps with sampling intervals matching the data generated in the next section. We discard the first 10000 points to remove transients. We use the next 10000 points to fit $\MB{y}^{t-d} = E\left[\mat{y}_{t-n}|\mat{x}_{t-1},\hdots,\mat{x}_{t-d}\right]$. We calculate the recursion error, 
\begin{equation}
\varepsilon_t = \mat{x}_{t}-\mat{F}\left( \mat{x}_{t-1},\mat{G}\left[ \mat{x}_{t-2},\hdots \mat{G}\left\{\mat{x}_{t-d},\MB{y}^{t-d}\right\} \hdots \right] \right)
\label{fullerror}
\end{equation}

We can also use a first order approximation to estimate the covariance, $\mat{\Sigma}_t$ of the apparent process noise $\varepsilon_t$, 

\begin{equation}
    \mat{\Sigma}_{t} \approx \J{P}{t-1} \J{Q}{t-2}\hdots \J{Q}{t-d}\mat{C}_{t-d} \J{Q}{t-d}^T\hdots \J{Q}{t-2}^T \J{P}{t-1}^T
\label{vard}
\end{equation}

\noindent where $\J{P}{t-1}$ is the matrix of partial derivatives of $\mat{F}$ with respect to $\mat{y}$ evaluated at $\mat{x}_{t-1}$ and $\mat{\bar{y}}^{t-1}$, $\J{Q}{t-n}$ is the matrix of partial derivatives of $\mat{G}$ with respect to $\mat{y}$ evaluated at $\mat{x}_{t-n}$ and $E\left[\mat{y}_{t-n}|\mat{x}_{t-1},\hdots,\mat{x}_{t-d}\right]$, and $\mat{C}_{t-d}$ is the covariance matrix for $\mat{y}_{t-d}$ conditional on $\mat{x}_{t-1},\hdots,\mat{x}_{t-d}$, i.e.
$\mat{C}_{t-d}=E\left[\left(\mat{y}-\MB{y}^{t-d}\right)\left(\mat{y}-\MB{y}^{t-d}\right)^T|\mat{x}_{t-1},\hdots,\mat{x}_{t-d}\right]$. 
 
Similar to the numerical estimation of the recursion error above, we can evaluate the first order approximation numerically from time series data by fitting $\mat{C}_{t-d}=E\left[\left(\mat{y}-\MB{y}^{t-d}\right)\left(\mat{y}-\MB{y}^{t-d}\right)^T|\mat{x}_{t-1},\hdots,\mat{x}_{t-d}\right]$ from time series data. The first order approximation is accurate for maps that are almost linear. For continuous time nonlinear dynamics, this would correspond to short sampling intervals.

Note that neither of these should be treated as strict bounds on practical accuracy. However, these can provide a baseline for the expected performance independent of the specifics of the forecast model.

\section{Recurrent Neural Network}
\label{rnn}
In a practical setting, the dynamics given by eq.~\eqref{two_states} can be learned from time-series data using delay vectors by fitting the function, \begin{align}
    \mat{x}_t = \mat{\hat{F}}\left(\mat{x}_{t-1},\hdots,\mat{x}_{t-d}\right)
\end{align}
This can be implemented directly using standard machine learning methods \cite{mandic2001recurrent,munch2017circumventing}. We implement a feedforward neural network (FNN) to approximate $\hat{\mat{F}}$ as a benchmark. The recursive form of eq.~\eqref{apxd1} suggests that the function approximator should be restricted among the space of functions that can be written as a recursive composition of lower dimensional functions $\mat{F}$ and $\mat{G}$. This can be achieved by constructing a Recurrent Neural Network (RNN) that imitates the recursive form in eq.~\eqref{apxd1},
\begin{align}
\mat{\hat{x}}_{t} &= \mat{W}_{\mat{x}} \mat{f}_{t} + \mat{b}_\mat{x} \nonumber \\ 
\mat{f}_{t} &= \mat{a}_\mat{f} \left(\mat{W}_{\mat{f}} (\mat{x}_{t-1} \oplus \mat{\hat{y}}_{t-1}) + \mat{b}_\mat{f}\right)\nonumber \\
\mat{\hat{y}}_{t-1} &= \mat{W}_{\mat{y}} \mat{g}_{t-1} + \mat{b}_\mat{y} \nonumber\\
\mat{g}_{t-1} &= \mat{a}_\mat{g}\left(\mat{W}_{\mat{g}} (\mat{x}_{t-2} \oplus  \mat{\hat{y}}_{t-2}) +\mat{b}_\mat{g}\right) \nonumber \\ 
&\vdots \nonumber \\
\mat{\hat{y}}_{t-d+1} &= \mat{W}_{\mat{y}} \mat{g}_{t-d+1}+\mat{b}_\mat{y} \nonumber \\ 
\mat{g}_{t-d+1} &= \mat{a}_\mat{g}\left(\mat{W}_{\mat{g}} (\mat{x}_{t-d} \oplus \mat{\hat{y}}_{t-d})+\mat{b}_\mat{g}\right)
\label{rnn_equations_1}
\end{align}

Where the functions $\mat{F}$ and $\mat{G}$ are approximated as single layer neural networks with hidden layer $\mat{f}$ and $\mat{g}$. $\mat{a}_\mat{f}$ and $\mat{a}_\mat{g}$ are the non-linear activation functions. In this study, $\mat{a}_\mat{f} = \mat{a}_\mat{g} = \tanh{}$. 
As $\mat{\hat{y}}_{t}$ is just a linear function of $\mat{g}_{t}$, it can be absorbed into the parameters $\mat{W}_{\mat{f}}$, $\mat{b}_\mat{f}$, $\mat{W}_{\mat{g}}$, and $\mat{b}_\mat{g}$ to obtain a simpler neural network,
\begin{align}
\mat{\hat{x}}_{t} &= \mat{W}_{\mat{x}} \mat{f}_{t} + \mat{b}_\mat{x} \nonumber \\ 
\mat{f}_{t} &= \mat{a}_\mat{f} \left(\mat{W}_{\mat{f}} (\mat{x}_{t-1} \oplus \mat{g}_{t-1}) + \mat{b}_\mat{f}\right)\nonumber \\
\mat{g}_{t-1} &= \mat{a}_\mat{g}\left(\mat{W}_{\mat{g}} (\mat{x}_{t-2} \oplus \mat{g}_{t-2})+\mat{b}_\mat{g}\right) \nonumber \\ 
&\vdots \nonumber \\
\mat{g}_{t-d+1} &= \mat{a}_\mat{g}\left(\mat{W}_{\mat{g}} (\mat{x}_{t-d} \oplus \mat{g}_{t-d})+\mat{b}_\mat{g}\right) 
\label{rnn_equations_2}
\end{align}
The model parameters, $\mat{W}_{\mat{\alpha}}$ and $\mat{b}_\mat{\alpha}$ are chosen to minimize the loss function,
\begin{align}
L = \sum_{t} \lVert\mat{\hat{x}}_t - \mat{x_t}^{\text{(data)}}\rVert^2
\end{align}
Note, since we don't observe $\mat{y}$, we can't compute $\mat{g}_{t-d}$. In this work, we choose $\mat{g}_{t-d}$ randomly for simplicity of setting up the backpropagation step. Alternatively, $g_{t-d}$ can be included in the training parameters. We train the parameters using backpropagation with RMSprop optimizer \cite{hinton2012neural}, and use early stopping \cite{prechelt1998automatic} to avoid over-fitting the training data. Note, since this is a proof-of-concept demonstration, we did not regularize using a penalty term in the loss function, as this would make it difficult to explicitly compare the FNN and RNN in terms of the NN complexity. In the following sections, we use simulated time series from popular nonlinear dynamics models namely, A) Discrete Lotka-Volterra model (2D), B) Lorenz 63 model \cite{lorenz1963deterministic} (3D), C) the Duffing oscillator \cite{duffing1918erzwungene} (4D), and D) the Lorenz 96 model \cite{lorenz1996predictability} (5D). In each of these cases, we use just the first variable to train the RNN (i.e., we only observe one variable). We train the RNN using training time-series of length $30$, $50$ and $100$ data points. We specifically focus on small training datasets as the advantage of the RNN over an FNN is larger in the data-poor regime. We expect this to be the case as the data rich cases will be equivalently fit with any function approximator, and systematic differences in performance would be hard to detect due to stochastic differences in training performance. We divide the training data further into `train' and `validation' sets that contain $75\%$ and $25\%$ of the data respectively for the datasets of size $50$ and $100$. The early stopping parameter is chosen to minimize validation loss. The errors reported are measured on out-of-sample `test' data of the same size as the training datasets. The errors were averaged across 100 different realizations of the model in each case.
\begin{table*}[]
    \centering
    \begin{tabular}{|c|c|c|c|c|}
        \hline
        Model & parameters & LE & Autocorrelation at $dt^{\text{a}}$ & RMSE `previous-value' predictor$^{\text{b}}$\\
        \hline
        Lotka Volterra & $r = [0.933,1.293]$, $A = [0.758,1.420]$ & $0.15$ & 0.632 & 0.858\\
        Lorenz63 & $\rho=28$,$\sigma=10$,$\beta=8/3$ & 0.91 & 0.869 & 0.512\\
        Duffing Oscillator & $[1.,-1.,0.3,0.5,1.2]$ & 0.17 & 0.667 & 0.816\\
        Lorenz96 & $N=5$, $F=8$ & $0.47$  & 0.866 & 0.518\\
        \hline
    \end{tabular}
    \caption{Descriptions of the datasets. a. Autocorrelation at the time-step used for predictions of the observed variable b. Normalized RMSE values from predicting using the previous value. Normalized such that using mean of the time-series leads to an RMSE$=1$.}
    \label{tab:model_params}
\end{table*}

\section{Numerical examples}
\label{illustrative_example}
\subsection{Discrete Lotka Volterra model}
\begin{figure*}
    \centering
    \includegraphics[width=0.9\textwidth]{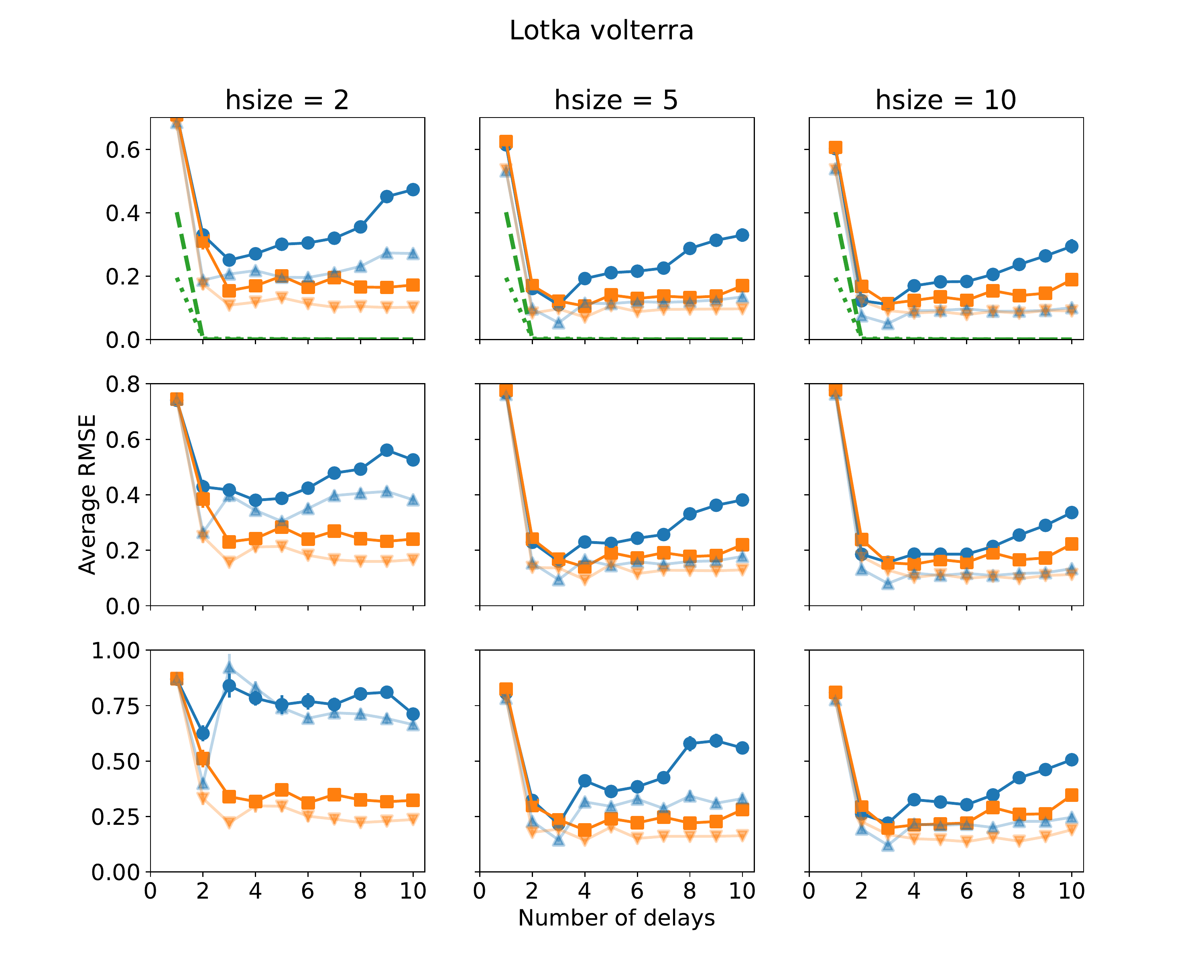}
    \caption{Normalized RMSE as a function of number of delays from a FNN (blue) vs RNN (orange) for the discrete Lotka-Volterra model \eqref{quadratic2d}. The top, middle, and bottom panels correspond to one-, two-, and three-steps ahead forecast error from a model trained on one-step ahead data. The left, middle, and right panels correspond to neural networks with hidden-layers with two, five and ten neurons each. The green dashed (dotted) lines in the top panel are numerical evaluations of the one-step ahead recursion error \eqref{apxd1} (and its first-order approximation \eqref{vard}). Dark (light) colors are results for with a training size of 50 (100) data points}
    \label{fig:results_lv}
\end{figure*}

To illustrate the effectiveness of the recursive approximation eq.~\eqref{apxd1}, we first examine a simple two-species system, where the state variables $x_t$ (observed) and $y_t$ (unobserved) are quadratic functions of $x_{t-1}$ and $y_{t-1}$. This is also known as a discrete Lotka-Volterra model in ecology literature \cite{lotka1910contribution}.
\begin{subequations}
\label{quadratic2d}
\begin{align}
    x_{t} = r_x x_{t-1}(1-x_{t-1}) + A_{xy} x_{t-1} y_{t-1}\label{quadratic2d-x}\\
    y_{t} = r_y y_{t-1}(1-y_{t-1}) + A_{yx} x_{t-1} y_{t-1}\label{quadratic2d-y}
\end{align}
\end{subequations}
Solving for $y_{t-1}$ using \eqref{quadratic2d-x} and substituting in \eqref{quadratic2d-y}, we get the evolution of $y_t$ as a function of $x$ alone,
\begin{equation}
    y_{t} = r_y \mathcal{Y}(x) \left(1 - \mathcal{Y}(x)\right) + A_{yx}x_{t-1}\mathcal{Y}(x)
    \label{quadratic2d-yofx}
\end{equation}
where,
\begin{equation*}
    \mathcal{Y}(x) = \left(\frac{x_t - r_x x_{t-1}(1-x_{t-1})}{A_{xy} x_{t-1}}\right)
\end{equation*}
Substituting \eqref{quadratic2d-yofx} back in \eqref{quadratic2d-x}, we obtain the dynamical equation only in terms of $x$

\begin{align}
    x_t &= r_x x_{t-1}(1-x_{t-1}) + \left[ r_y\left(\frac{x_{t-1}^2}{x_{t-2}} - r_x x_{t-1} (1-x_{t-2})\right)\right.\nonumber\\
    &\hspace{3cm}\times\left.\left(1-\frac{x_{t-1}-r_x x_{t-2}(1-x_{t-2})}{A_{xy} x_{t-2}}\right)\right. \nonumber\\
    &\hspace{1cm} + A_{yx} x_{t-1} (x_{t-1} - r_x x_{t-2}(1-x_{t-2})) \bigg]
    \label{quadratic-delay-exact}
\end{align}
which is highly nonlinear. Note the resulting single-variable dynamics is no longer quadratic and has arbitrarily higher-order non-zero derivatives. However the recursive approximation \eqref{apxd1} has a simpler functional form and ensures that higher derivatives are bounded. Specifically, 
\begin{align}
    x_t &\approx r_x x_{t-1}(1-x_{t-1}) + \nonumber\\
    &\quad A_{xy} x_{t-1}\left[r_y y_{t-2}(1-y^*_{t-2}) + A_{yx} x_{t-2} y^*_{t-2}\right]
\label{quadratic-recursive1}
\end{align}
has non-zero derivatives only up to second order in x (and in general up to order $d$) considering $y_t$s are constants, thus requiring less data to reconstruct the approximate dynamics. We calculate the theoretical recursion error $\epsilon$ given by Eq.~\eqref{fullerror} and its first order approximation, Eq.~\eqref{vard} in the limit of large data. We see that the errors go to zero when the number of delays is two or greater consistent with Eq.~\eqref{quadratic-delay-exact}.
We generate time-series of length $50$ and $100$ data points and compare the performance of the RNN architecture \eqref{rnn_equations_2} vs. FNN across different hidden layer sizes. The hidden layer size limits the expressivity of a neural network. For example, a neural network with one-dimensional input and two hidden neurons can only fit a function with a single peak. 
We see a significant improvement in performance of RNN over FNN for small hidden layer sizes as expected due to the simpler functions $\mat{F}$ and $\mat{G}$ required to be fit by the RNN (see Fig.~\ref{fig:results_lv}) as against the more complex delay function $\mat{\tilde{F}}$. The small number of hidden neurons forces the neural networks to fit a function with less features, thereby making it difficult to fit the highly nonlinear function in \eqref{quadratic-delay-exact}. We also see that the difference in performance between the RNN and FNN widens with increasing number of delays due to the increasing complexity  $\mat{\tilde{F}}$. We also compare multi-step ahead predictions using the single-step neural networks and iteratively applying the function on the delay vector to produce the next state. We hypothesize that the RNN functions $\mat{f}$ and $\mat{g}$ should be better in producing the multi-step forecast. This is because the iteration of the more complex $\mat{\hat{F}}$ can lead to a larger variance at the locations in state space not seen by the one-step training data, compared to iterating the lower dimensional function $\mat{f}$. We see that the RNN indeed performs better than FNN in two- and three-step ahead prediction. The FNN errors increase significantly more than the RNN as we increase the number of steps hinting at the robustness of the recurrent structure of RNN for dynamical systems prediction.

We next look at some popular continuous chaotic dynamics in higher dimensions.

\subsection{Lorenz 63 model}
Lorenz 63 is one of the most popular chaotic models. It is a first order differential equation modeling a simplified version of atmospheric convection \cite{lorenz1963deterministic}.
\begin{align}
    \dot{x} &= \sigma (y-x) \nonumber \\
    \dot{y} &= x(\rho-z)-y \nonumber \\
    \dot{z} &= x y - \beta z
    \label{lorenz63}
\end{align}
We chose a sampling rate of $10$ Hz so that the prediction problem was sufficiently non-trivial, but not impossible (see Table ~\ref{tab:model_params} for details). The observed variable is $x$. We compute training and validation loss for the neural networks with 2-20 hidden neurons (same number of hidden neurons for both $\mat{f}$ and $\mat{g}$ in case of RNN), and choose the one with the minimum validation loss. We also use the validation loss for `early stopping' the training. The recursion error \eqref{fullerror}, and its approximation \eqref{vard} tend to zero with three or more delays.

The optimal number of delays for both the FNN and the RNN is three (see Fig.~\ref{fig:results}). The optimal RMSE is statistically indistinguishable between the FNN and the RNN. However, the RNN has a robust performance across all number of delays (especially for the smaller dataset), which may be desirable for automated applications. 

We also plot the hidden-layer size-specific results (see Appendix Fig.~\ref{fig:results_l63}). We see a similar trend as the discrete Lotka-Volterra model for the smallest hidden layer size ($h=2$), but there is no systematic advantage to RNN with larger hidden layers.

\subsection{Duffing oscillator}

The Duffing oscillator is a second order differential equation with periodic forcing,
\begin{equation}
    \Ddot{x} + \delta \dot{x} + \beta x + \alpha x^3 = \gamma \cos{(\omega t)}
\end{equation}
This can be re-written as a first-order autonomous system by introducing the variables $y = \dot{x}$, $v = \cos{(\omega t)}$ and $z = \sin{(\omega t)}$,
\begin{align}
    \dot{x} &= y \nonumber\\
    \dot{y} &= \gamma v - \delta y - \beta x - \alpha x^3 \nonumber\\
    \dot{v} &= -\omega z \nonumber \\
    \dot{z} &= \omega v
    \label{duffing}
\end{align}
We chose a sampling rate of $1$ Hz, as this model has a Lyapunov horizon that is roughly an order of magnitude larger than the Lorenz63 model (see Table ~\ref{tab:model_params}). The observed variable is $x$. The recursion error \eqref{fullerror}, and is approximation \eqref{vard} go to zero with four or more delays. The optimal number of delays in this case is four (see Fig.~\ref{fig:results}). The results for this model are qualitatively similar to the Lorenz 63 model, that is, at optimal number of delays, the performance is indistinguishable for the two neural networks, but RNN is more robust across the number of delays. This trend is similar even when we restrict the neural networks to have small hidden layers (see Appendix Fig.~\ref{fig:results_df}).

\subsection{Lorenz 96 model}
\begin{figure*}[ht]
\centering
\includegraphics[width=0.9\textwidth]{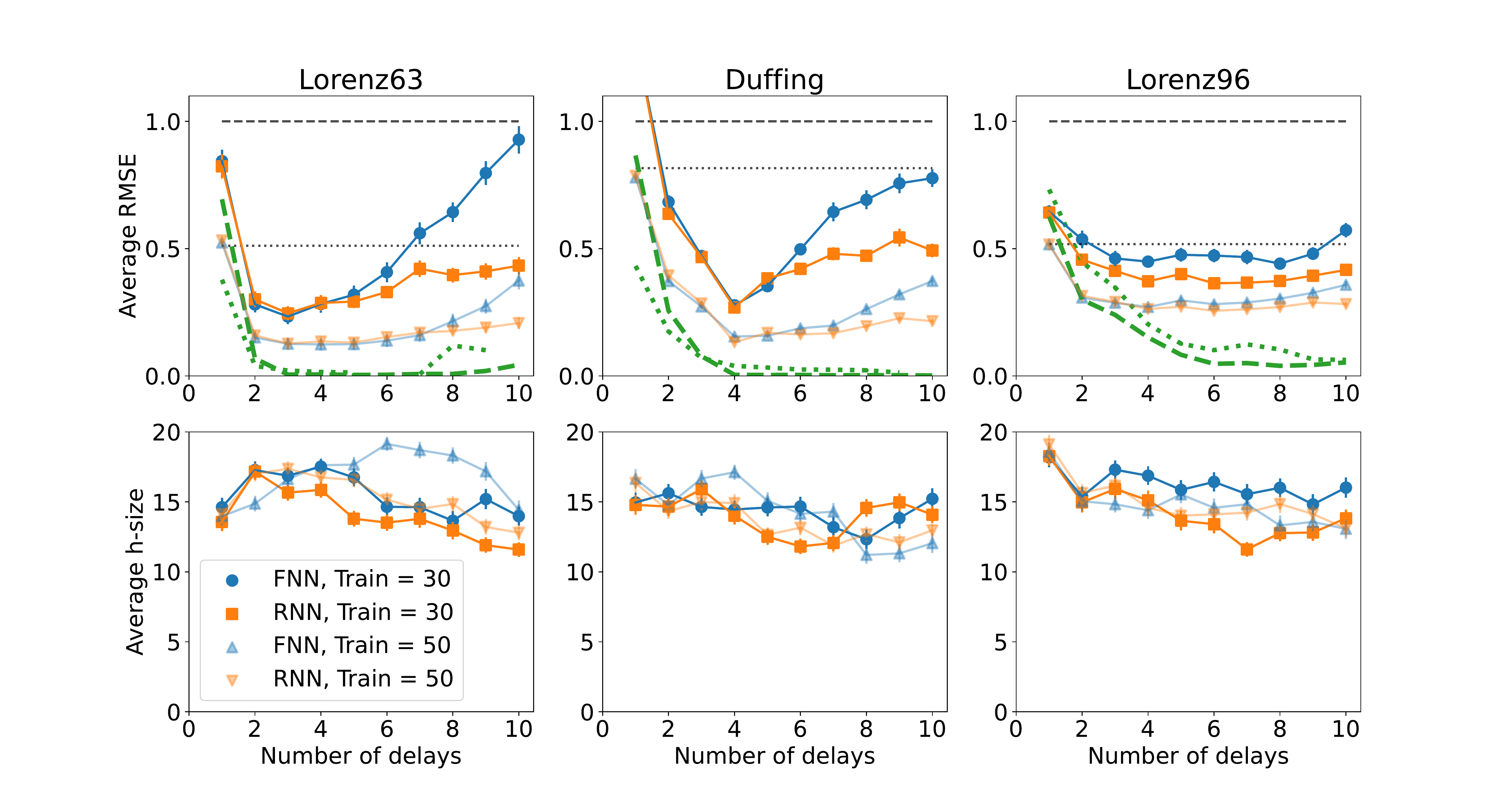}
\caption{(Top) Normalized RMSE as a function of number of delays from a FNN (blue) vs RNN (orange) for the Lorenz 63 model (left), Duffing (middle), and Lorenz 96 model in 5D (right) with parameters in Table~\ref{tab:model_params}. The black dashed (dotted) lines are benchmark predictions using the time series mean (previous value). The green dashed (dotted) lines are numerical evaluations of the recursion error \eqref{apxd1} (and its first-order approximation \eqref{vard}). (Bottom) The average number of optimal hidden-neurons. Dark (light) colors are results for with a training size of 30 (50) data points}
\label{fig:results}
\end{figure*}

Lorenz 96 is a popular model to test tools for chaotic time series prediction in a high dimensional setting \cite{chattopadhyay2020datadriven,dueben2018challenges}.
We generated time series data using the Lorenz 96 model \cite{lorenz1996predictability}
\begin{align}
    \label{lorenz96}
    \frac{\mathrm{d} x_i}{\mathrm{d} t} = \left(x_{i+1} - x_{i-2}\right) x_{i-1} - x_i + F,\quad 1\le i \le N
\end{align}
where it is assumed that $x_{-1}=x_{N-1}$, $x_{0}=x_{N}$ and $x_{N+1}=x_{1}$. We use the parameters $N=5$, $F=8$. The dynamics are chaotic with Lyapunov exponent $= 0.472 \pm 0.002$. Sampling rate is $10$ Hz. The observed variable is $x_1$. The recursion error \eqref{fullerror}, and is approximation \eqref{vard} tends to zero with roughly six or more delays. The optimal number of delays is six and eight for the RNN and FNN respectively (see Fig.~\ref{fig:results}).

The RNN shows significantly better performance as measured by the average RMSE in the case of the smaller dataset. There is also a significant difference between the optimal number of hidden neurons between the two NNs, with the RNN opting for lower number of hidden neurons indicating that the function to be fit is of a lower complexity. However there is no significant difference in performance when the neural networks are restricted to small sizes (see Appendix Fig.~\ref{fig:results_l96}).

\section{Discussion}
In this study, we explore the possibility that choosing the neural network architecture that is derived from the structure of generating dynamics can lead to more efficient recovery of the dynamics from data. We see evidence for this by a significantly better prediction performance by the Recurrent Neural Network as compared to the Feedforward Neural Network in the small-data regime for the discrete Lotka-Volterra model. We also see that the RNN increasingly outperforms FNN in multistep prediction tasks when the dynamics are trained on singlestep data. The systematic advantage RNN has over FNN when trained with small hidden layers suggest that the smoothness of $\mat{F}$ and $\mat{G}$ functions compared to the delay vector $\mat{\tilde{F}}$ can be leveraged by manipulating the structure of neural networks.

In this study, we see that the structural advantage of RNN comes into play when the attractor dimension is small and the manifold is smooth. The advantage of RNN seems to systematically diminish with increasing dimensionality of dynamics. This is consistent with the interpretation that the RNN is exposed to smoother functional forms than the FNN which is directly fit to the more nonlinear delay map. More work needs to be done to fully characterize the regime where incorporating the dynamical structure in the neural network will yield better predictions.

Having dynamically meaningful units within the neural networks is useful in applications where it is important to learn the mechanistic relationships between variables. It also makes incorporating auxiliary information straightforward. For example, information about interactions between states can be implemented by conditioning the model to constrain the partial derivatives $\partial{H}_i /\partial{z}_j = 0$ (where $H \in \{F, G\}$
and $z \in \{x, y\}$) that correspond to non-interacting states. This can be achieved through regularization or constraint optimization of neural networks. Our methods can be extended to take advantage of the structure of spatial dynamics as well. Since we expect the spatial interaction structure to be sparse, we expect $\mat{F}$ and $\mat{G}$ to have a much lower dimensionality compared to fitting the full delay-embedding function.

In the field of statistical mechanics, the problem of unobserved states has been addressed by the Mori Zwanzig formulation where the Zwanzig operator is used to project the dynamics onto the linear subspace of the observed dynamics, where the ignored degrees of freedom appear as a memory term and a noise term. Calculating the memory term for nonlinear dynamics is non-trivial, and requires the expansion of the basis to lift the dynamics to a linear space. This can lead to an unbounded expansion of the state space in chaotic systems. In contrast, our approximation provides a straightforward way to incorporate the induced memory from partial observations.

In summary, we address the gap in the theoretical literature on the efficacy of recurrent neural networks. We show how partially observed dynamics can be restructured to reveal a recurrent structure, which can be learnt by fitting recurrent neural networks on time series data. We also provide a connection to time-delay embedding and discuss the potential applications of this methodology.

This work was supported by NOAA's HPCC incubator.

\bibliography{refs}

\begin{thebibliography}{36}%
\makeatletter
\providecommand \@ifxundefined [1]{%
 \@ifx{#1\undefined}
}%
\providecommand \@ifnum [1]{%
 \ifnum #1\expandafter \@firstoftwo
 \else \expandafter \@secondoftwo
 \fi
}%
\providecommand \@ifx [1]{%
 \ifx #1\expandafter \@firstoftwo
 \else \expandafter \@secondoftwo
 \fi
}%
\providecommand \natexlab [1]{#1}%
\providecommand \enquote  [1]{``#1''}%
\providecommand \bibnamefont  [1]{#1}%
\providecommand \bibfnamefont [1]{#1}%
\providecommand \citenamefont [1]{#1}%
\providecommand \href@noop [0]{\@secondoftwo}%
\providecommand \href [0]{\begingroup \@sanitize@url \@href}%
\providecommand \@href[1]{\@@startlink{#1}\@@href}%
\providecommand \@@href[1]{\endgroup#1\@@endlink}%
\providecommand \@sanitize@url [0]{\catcode `\\12\catcode `\$12\catcode
  `\&12\catcode `\#12\catcode `\^12\catcode `\_12\catcode `\%12\relax}%
\providecommand \@@startlink[1]{}%
\providecommand \@@endlink[0]{}%
\providecommand \url  [0]{\begingroup\@sanitize@url \@url }%
\providecommand \@url [1]{\endgroup\@href {#1}{\urlprefix }}%
\providecommand \urlprefix  [0]{URL }%
\providecommand \Eprint [0]{\href }%
\providecommand \doibase [0]{https://doi.org/}%
\providecommand \selectlanguage [0]{\@gobble}%
\providecommand \bibinfo  [0]{\@secondoftwo}%
\providecommand \bibfield  [0]{\@secondoftwo}%
\providecommand \translation [1]{[#1]}%
\providecommand \BibitemOpen [0]{}%
\providecommand \bibitemStop [0]{}%
\providecommand \bibitemNoStop [0]{.\EOS\space}%
\providecommand \EOS [0]{\spacefactor3000\relax}%
\providecommand \BibitemShut  [1]{\csname bibitem#1\endcsname}%
\let\auto@bib@innerbib\@empty
\bibitem [{\citenamefont {Palmer}(2001)}]{palmer2001nonlinear}%
  \BibitemOpen
  \bibfield  {author} {\bibinfo {author} {\bibfnamefont {T.~N.}\ \bibnamefont
  {Palmer}},\ }\bibfield  {title} {\bibinfo {title} {A nonlinear dynamical
  perspective on model error: A proposal for non-local stochastic-dynamic
  parametrization in weather and climate prediction models},\ }\href@noop {}
  {\bibfield  {journal} {\bibinfo  {journal} {Quarterly Journal of the Royal
  Meteorological Society}\ }\textbf {\bibinfo {volume} {127}},\ \bibinfo
  {pages} {279} (\bibinfo {year} {2001})}\BibitemShut {NoStop}%
\bibitem [{\citenamefont {Lek}\ \emph {et~al.}(1996)\citenamefont {Lek},
  \citenamefont {Delacoste}, \citenamefont {Baran}, \citenamefont {Dimopoulos},
  \citenamefont {Lauga},\ and\ \citenamefont {Aulagnier}}]{lek1996application}%
  \BibitemOpen
  \bibfield  {author} {\bibinfo {author} {\bibfnamefont {S.}~\bibnamefont
  {Lek}}, \bibinfo {author} {\bibfnamefont {M.}~\bibnamefont {Delacoste}},
  \bibinfo {author} {\bibfnamefont {P.}~\bibnamefont {Baran}}, \bibinfo
  {author} {\bibfnamefont {I.}~\bibnamefont {Dimopoulos}}, \bibinfo {author}
  {\bibfnamefont {J.}~\bibnamefont {Lauga}},\ and\ \bibinfo {author}
  {\bibfnamefont {S.}~\bibnamefont {Aulagnier}},\ }\bibfield  {title} {\bibinfo
  {title} {Application of neural networks to modelling nonlinear relationships
  in ecology},\ }\href@noop {} {\bibfield  {journal} {\bibinfo  {journal}
  {Ecological modelling}\ }\textbf {\bibinfo {volume} {90}},\ \bibinfo {pages}
  {39} (\bibinfo {year} {1996})}\BibitemShut {NoStop}%
\bibitem [{\citenamefont {Luo}\ \emph {et~al.}(2011)\citenamefont {Luo},
  \citenamefont {Ogle}, \citenamefont {Tucker}, \citenamefont {Fei},
  \citenamefont {Gao}, \citenamefont {LaDeau}, \citenamefont {Clark},\ and\
  \citenamefont {Schimel}}]{luo2011ecological}%
  \BibitemOpen
  \bibfield  {author} {\bibinfo {author} {\bibfnamefont {Y.}~\bibnamefont
  {Luo}}, \bibinfo {author} {\bibfnamefont {K.}~\bibnamefont {Ogle}}, \bibinfo
  {author} {\bibfnamefont {C.}~\bibnamefont {Tucker}}, \bibinfo {author}
  {\bibfnamefont {S.}~\bibnamefont {Fei}}, \bibinfo {author} {\bibfnamefont
  {C.}~\bibnamefont {Gao}}, \bibinfo {author} {\bibfnamefont {S.}~\bibnamefont
  {LaDeau}}, \bibinfo {author} {\bibfnamefont {J.~S.}\ \bibnamefont {Clark}},\
  and\ \bibinfo {author} {\bibfnamefont {D.~S.}\ \bibnamefont {Schimel}},\
  }\bibfield  {title} {\bibinfo {title} {Ecological forecasting and data
  assimilation in a data-rich era},\ }\href@noop {} {\bibfield  {journal}
  {\bibinfo  {journal} {Ecological Applications}\ }\textbf {\bibinfo {volume}
  {21}},\ \bibinfo {pages} {1429} (\bibinfo {year} {2011})}\BibitemShut
  {NoStop}%
\bibitem [{\citenamefont {Lachowicz}(2011)}]{lachowicz2011individually}%
  \BibitemOpen
  \bibfield  {author} {\bibinfo {author} {\bibfnamefont {M.}~\bibnamefont
  {Lachowicz}},\ }\bibfield  {title} {\bibinfo {title} {Individually-based
  markov processes modeling nonlinear systems in mathematical biology},\
  }\href@noop {} {\bibfield  {journal} {\bibinfo  {journal} {Nonlinear
  Analysis: Real World Applications}\ }\textbf {\bibinfo {volume} {12}},\
  \bibinfo {pages} {2396} (\bibinfo {year} {2011})}\BibitemShut {NoStop}%
\bibitem [{\citenamefont {Imoto}\ \emph {et~al.}(2003)\citenamefont {Imoto},
  \citenamefont {Kim}, \citenamefont {Goto}, \citenamefont {Aburatani},
  \citenamefont {Tashiro}, \citenamefont {Kuhara},\ and\ \citenamefont
  {Miyano}}]{imoto2003bayesian}%
  \BibitemOpen
  \bibfield  {author} {\bibinfo {author} {\bibfnamefont {S.}~\bibnamefont
  {Imoto}}, \bibinfo {author} {\bibfnamefont {S.}~\bibnamefont {Kim}}, \bibinfo
  {author} {\bibfnamefont {T.}~\bibnamefont {Goto}}, \bibinfo {author}
  {\bibfnamefont {S.}~\bibnamefont {Aburatani}}, \bibinfo {author}
  {\bibfnamefont {K.}~\bibnamefont {Tashiro}}, \bibinfo {author} {\bibfnamefont
  {S.}~\bibnamefont {Kuhara}},\ and\ \bibinfo {author} {\bibfnamefont
  {S.}~\bibnamefont {Miyano}},\ }\bibfield  {title} {\bibinfo {title} {Bayesian
  network and nonparametric heteroscedastic regression for nonlinear modeling
  of genetic network},\ }\href@noop {} {\bibfield  {journal} {\bibinfo
  {journal} {Journal of bioinformatics and computational biology}\ }\textbf
  {\bibinfo {volume} {1}},\ \bibinfo {pages} {231} (\bibinfo {year}
  {2003})}\BibitemShut {NoStop}%
\bibitem [{\citenamefont {Farmer}\ and\ \citenamefont
  {Sidorowich}(1987)}]{farmer1987predicting}%
  \BibitemOpen
  \bibfield  {author} {\bibinfo {author} {\bibfnamefont {J.~D.}\ \bibnamefont
  {Farmer}}\ and\ \bibinfo {author} {\bibfnamefont {J.~J.}\ \bibnamefont
  {Sidorowich}},\ }\bibfield  {title} {\bibinfo {title} {Predicting chaotic
  time series},\ }\href@noop {} {\bibfield  {journal} {\bibinfo  {journal}
  {Physical review letters}\ }\textbf {\bibinfo {volume} {59}},\ \bibinfo
  {pages} {845} (\bibinfo {year} {1987})}\BibitemShut {NoStop}%
\bibitem [{\citenamefont {Takens}(1981)}]{takens1981detecting}%
  \BibitemOpen
  \bibfield  {author} {\bibinfo {author} {\bibfnamefont {F.}~\bibnamefont
  {Takens}},\ }\bibfield  {title} {\bibinfo {title} {Detecting strange
  attractors in turbulence},\ }in\ \href@noop {} {\emph {\bibinfo {booktitle}
  {Dynamical systems and turbulence, Warwick 1980}}}\ (\bibinfo  {publisher}
  {Springer},\ \bibinfo {year} {1981})\ pp.\ \bibinfo {pages}
  {366--381}\BibitemShut {NoStop}%
\bibitem [{\citenamefont {Sugihara}\ and\ \citenamefont
  {May}(1990)}]{sugihara1990nonlinear}%
  \BibitemOpen
  \bibfield  {author} {\bibinfo {author} {\bibfnamefont {G.}~\bibnamefont
  {Sugihara}}\ and\ \bibinfo {author} {\bibfnamefont {R.~M.}\ \bibnamefont
  {May}},\ }\bibfield  {title} {\bibinfo {title} {Nonlinear forecasting as a
  way of distinguishing chaos from measurement error in time series},\
  }\href@noop {} {\bibfield  {journal} {\bibinfo  {journal} {Nature}\ }\textbf
  {\bibinfo {volume} {344}},\ \bibinfo {pages} {734} (\bibinfo {year}
  {1990})}\BibitemShut {NoStop}%
\bibitem [{\citenamefont {Ragwitz}\ and\ \citenamefont
  {Kantz}(2002)}]{ragwitz2002markov}%
  \BibitemOpen
  \bibfield  {author} {\bibinfo {author} {\bibfnamefont {M.}~\bibnamefont
  {Ragwitz}}\ and\ \bibinfo {author} {\bibfnamefont {H.}~\bibnamefont
  {Kantz}},\ }\bibfield  {title} {\bibinfo {title} {Markov models from data by
  simple nonlinear time series predictors in delay embedding spaces},\
  }\href@noop {} {\bibfield  {journal} {\bibinfo  {journal} {Physical Review
  E}\ }\textbf {\bibinfo {volume} {65}},\ \bibinfo {pages} {056201} (\bibinfo
  {year} {2002})}\BibitemShut {NoStop}%
\bibitem [{\citenamefont {Garcia}\ and\ \citenamefont
  {Almeida}(2005)}]{garcia2005multivariate}%
  \BibitemOpen
  \bibfield  {author} {\bibinfo {author} {\bibfnamefont {S.~P.}\ \bibnamefont
  {Garcia}}\ and\ \bibinfo {author} {\bibfnamefont {J.~S.}\ \bibnamefont
  {Almeida}},\ }\bibfield  {title} {\bibinfo {title} {Multivariate phase space
  reconstruction by nearest neighbor embedding with different time delays},\
  }\href@noop {} {\bibfield  {journal} {\bibinfo  {journal} {Physical Review
  E}\ }\textbf {\bibinfo {volume} {72}},\ \bibinfo {pages} {027205} (\bibinfo
  {year} {2005})}\BibitemShut {NoStop}%
\bibitem [{\citenamefont {Qin}\ \emph {et~al.}(2019)\citenamefont {Qin},
  \citenamefont {Wu},\ and\ \citenamefont {Xiu}}]{qin2019data}%
  \BibitemOpen
  \bibfield  {author} {\bibinfo {author} {\bibfnamefont {T.}~\bibnamefont
  {Qin}}, \bibinfo {author} {\bibfnamefont {K.}~\bibnamefont {Wu}},\ and\
  \bibinfo {author} {\bibfnamefont {D.}~\bibnamefont {Xiu}},\ }\bibfield
  {title} {\bibinfo {title} {Data driven governing equations approximation
  using deep neural networks},\ }\href@noop {} {\bibfield  {journal} {\bibinfo
  {journal} {Journal of Computational Physics}\ }\textbf {\bibinfo {volume}
  {395}},\ \bibinfo {pages} {620} (\bibinfo {year} {2019})}\BibitemShut
  {NoStop}%
\bibitem [{\citenamefont {Raissi}\ \emph {et~al.}(2018)\citenamefont {Raissi},
  \citenamefont {Perdikaris},\ and\ \citenamefont
  {Karniadakis}}]{raissi2018multistep}%
  \BibitemOpen
  \bibfield  {author} {\bibinfo {author} {\bibfnamefont {M.}~\bibnamefont
  {Raissi}}, \bibinfo {author} {\bibfnamefont {P.}~\bibnamefont {Perdikaris}},\
  and\ \bibinfo {author} {\bibfnamefont {G.~E.}\ \bibnamefont {Karniadakis}},\
  }\bibfield  {title} {\bibinfo {title} {Multistep neural networks for
  data-driven discovery of nonlinear dynamical systems},\ }\href@noop {}
  {\bibfield  {journal} {\bibinfo  {journal} {arXiv preprint arXiv:1801.01236}\
  } (\bibinfo {year} {2018})}\BibitemShut {NoStop}%
\bibitem [{\citenamefont {Lusch}\ \emph {et~al.}(2018)\citenamefont {Lusch},
  \citenamefont {Kutz},\ and\ \citenamefont {Brunton}}]{lusch2018deep}%
  \BibitemOpen
  \bibfield  {author} {\bibinfo {author} {\bibfnamefont {B.}~\bibnamefont
  {Lusch}}, \bibinfo {author} {\bibfnamefont {J.~N.}\ \bibnamefont {Kutz}},\
  and\ \bibinfo {author} {\bibfnamefont {S.~L.}\ \bibnamefont {Brunton}},\
  }\bibfield  {title} {\bibinfo {title} {Deep learning for universal linear
  embeddings of nonlinear dynamics},\ }\href@noop {} {\bibfield  {journal}
  {\bibinfo  {journal} {Nature communications}\ }\textbf {\bibinfo {volume}
  {9}},\ \bibinfo {pages} {1} (\bibinfo {year} {2018})}\BibitemShut {NoStop}%
\bibitem [{\citenamefont {Weinan}\ \emph {et~al.}(2020)\citenamefont {Weinan},
  \citenamefont {Ma}, \citenamefont {Wojtowytsch},\ and\ \citenamefont
  {Wu}}]{weinan2020towards}%
  \BibitemOpen
  \bibfield  {author} {\bibinfo {author} {\bibfnamefont {E.}~\bibnamefont
  {Weinan}}, \bibinfo {author} {\bibfnamefont {C.}~\bibnamefont {Ma}}, \bibinfo
  {author} {\bibfnamefont {S.}~\bibnamefont {Wojtowytsch}},\ and\ \bibinfo
  {author} {\bibfnamefont {L.}~\bibnamefont {Wu}},\ }\bibfield  {title}
  {\bibinfo {title} {Towards a mathematical understanding of neural
  network-based machine learning: What we know and what we don’t},\
  }\href@noop {} {\bibfield  {journal} {\bibinfo  {journal} {arXiv preprint
  arXiv:2009.10713}\ } (\bibinfo {year} {2020})}\BibitemShut {NoStop}%
\bibitem [{\citenamefont {Caterini}\ and\ \citenamefont
  {Chang}(2018)}]{caterini2018deep}%
  \BibitemOpen
  \bibfield  {author} {\bibinfo {author} {\bibfnamefont {A.~L.}\ \bibnamefont
  {Caterini}}\ and\ \bibinfo {author} {\bibfnamefont {D.~E.}\ \bibnamefont
  {Chang}},\ }\href@noop {} {\emph {\bibinfo {title} {Deep Neural Networks in a
  Mathematical Framework}}}\ (\bibinfo  {publisher} {Springer},\ \bibinfo
  {year} {2018})\BibitemShut {NoStop}%
\bibitem [{\citenamefont {Mandic}\ and\ \citenamefont
  {Chambers}(2001)}]{mandic2001recurrent}%
  \BibitemOpen
  \bibfield  {author} {\bibinfo {author} {\bibfnamefont {D.}~\bibnamefont
  {Mandic}}\ and\ \bibinfo {author} {\bibfnamefont {J.}~\bibnamefont
  {Chambers}},\ }\href@noop {} {\emph {\bibinfo {title} {Recurrent neural
  networks for prediction: learning algorithms, architectures and stability}}}\
  (\bibinfo  {publisher} {Wiley},\ \bibinfo {year} {2001})\BibitemShut
  {NoStop}%
\bibitem [{\citenamefont {Connor}\ \emph {et~al.}(1994)\citenamefont {Connor},
  \citenamefont {Martin},\ and\ \citenamefont {Atlas}}]{connor1994recurrent}%
  \BibitemOpen
  \bibfield  {author} {\bibinfo {author} {\bibfnamefont {J.}~\bibnamefont
  {Connor}}, \bibinfo {author} {\bibfnamefont {R.}~\bibnamefont {Martin}},\
  and\ \bibinfo {author} {\bibfnamefont {L.}~\bibnamefont {Atlas}},\ }\bibfield
   {title} {\bibinfo {title} {Recurrent neural networks and robust time series
  prediction},\ }\href {https://doi.org/10.1109/72.279188} {\bibfield
  {journal} {\bibinfo  {journal} {IEEE Transactions on Neural Networks}\
  }\textbf {\bibinfo {volume} {5}},\ \bibinfo {pages} {240} (\bibinfo {year}
  {1994})}\BibitemShut {NoStop}%
\bibitem [{\citenamefont {Goodfellow}\ \emph {et~al.}(2016)\citenamefont
  {Goodfellow}, \citenamefont {Bengio},\ and\ \citenamefont
  {Courville}}]{goodfellow2016deep}%
  \BibitemOpen
  \bibfield  {author} {\bibinfo {author} {\bibfnamefont {I.}~\bibnamefont
  {Goodfellow}}, \bibinfo {author} {\bibfnamefont {Y.}~\bibnamefont {Bengio}},\
  and\ \bibinfo {author} {\bibfnamefont {A.}~\bibnamefont {Courville}},\
  }\href@noop {} {\emph {\bibinfo {title} {Deep learning}}}\ (\bibinfo
  {publisher} {MIT press},\ \bibinfo {year} {2016})\BibitemShut {NoStop}%
\bibitem [{\citenamefont {Williams}\ and\ \citenamefont
  {Zipser}(1989)}]{williams1989learning}%
  \BibitemOpen
  \bibfield  {author} {\bibinfo {author} {\bibfnamefont {R.~J.}\ \bibnamefont
  {Williams}}\ and\ \bibinfo {author} {\bibfnamefont {D.}~\bibnamefont
  {Zipser}},\ }\bibfield  {title} {\bibinfo {title} {{A Learning Algorithm for
  Continually Running Fully Recurrent Neural Networks}},\ }\href
  {https://doi.org/10.1162/neco.1989.1.2.270} {\bibfield  {journal} {\bibinfo
  {journal} {Neural Computation}\ }\textbf {\bibinfo {volume} {1}},\ \bibinfo
  {pages} {270} (\bibinfo {year} {1989})},\ \Eprint
  {https://arxiv.org/abs/https://direct.mit.edu/neco/article-pdf/1/2/270/811849/neco.1989.1.2.270.pdf}
  {https://direct.mit.edu/neco/article-pdf/1/2/270/811849/neco.1989.1.2.270.pdf}
  \BibitemShut {NoStop}%
\bibitem [{\citenamefont {Cheng}\ and\ \citenamefont
  {Tong}(1994)}]{cheng1994orthogonal}%
  \BibitemOpen
  \bibfield  {author} {\bibinfo {author} {\bibfnamefont {B.}~\bibnamefont
  {Cheng}}\ and\ \bibinfo {author} {\bibfnamefont {H.}~\bibnamefont {Tong}},\
  }\bibfield  {title} {\bibinfo {title} {Orthogonal projection, embedding
  dimension and sample size in chaotic time series from a statistical
  perspective},\ }\href@noop {} {\bibfield  {journal} {\bibinfo  {journal}
  {Philosophical Transactions of the Royal Society of London. Series A:
  Physical and Engineering Sciences}\ }\textbf {\bibinfo {volume} {348}},\
  \bibinfo {pages} {325} (\bibinfo {year} {1994})}\BibitemShut {NoStop}%
\bibitem [{\citenamefont {Munch}\ \emph {et~al.}(2017)\citenamefont {Munch},
  \citenamefont {Poynor},\ and\ \citenamefont
  {Arriaza}}]{munch2017circumventing}%
  \BibitemOpen
  \bibfield  {author} {\bibinfo {author} {\bibfnamefont {S.~B.}\ \bibnamefont
  {Munch}}, \bibinfo {author} {\bibfnamefont {V.}~\bibnamefont {Poynor}},\ and\
  \bibinfo {author} {\bibfnamefont {J.~L.}\ \bibnamefont {Arriaza}},\
  }\bibfield  {title} {\bibinfo {title} {Circumventing structural uncertainty:
  A bayesian perspective on nonlinear forecasting for ecology},\ }\href
  {https://doi.org/https://doi.org/10.1016/j.ecocom.2016.08.006} {\bibfield
  {journal} {\bibinfo  {journal} {Ecological Complexity}\ }\textbf {\bibinfo
  {volume} {32}},\ \bibinfo {pages} {134 } (\bibinfo {year} {2017})},\ \bibinfo
  {note} {uncertainty in Ecology}\BibitemShut {NoStop}%
\bibitem [{\citenamefont {Kantz}\ and\ \citenamefont
  {Olbrich}(1997)}]{kantz1997scalar}%
  \BibitemOpen
  \bibfield  {author} {\bibinfo {author} {\bibfnamefont {H.}~\bibnamefont
  {Kantz}}\ and\ \bibinfo {author} {\bibfnamefont {E.}~\bibnamefont
  {Olbrich}},\ }\bibfield  {title} {\bibinfo {title} {Scalar observations from
  a class of high-dimensional chaotic systems: Limitations of the time delay
  embedding},\ }\href@noop {} {\bibfield  {journal} {\bibinfo  {journal}
  {Chaos: An Interdisciplinary Journal of Nonlinear Science}\ }\textbf
  {\bibinfo {volume} {7}},\ \bibinfo {pages} {423} (\bibinfo {year}
  {1997})}\BibitemShut {NoStop}%
\bibitem [{\citenamefont {Elman}(1990)}]{elman1990finding}%
  \BibitemOpen
  \bibfield  {author} {\bibinfo {author} {\bibfnamefont {J.~L.}\ \bibnamefont
  {Elman}},\ }\bibfield  {title} {\bibinfo {title} {Finding structure in
  time},\ }\href@noop {} {\bibfield  {journal} {\bibinfo  {journal} {Cognitive
  science}\ }\textbf {\bibinfo {volume} {14}},\ \bibinfo {pages} {179}
  (\bibinfo {year} {1990})}\BibitemShut {NoStop}%
\bibitem [{\citenamefont {Han}\ \emph {et~al.}(2004)\citenamefont {Han},
  \citenamefont {Shi},\ and\ \citenamefont {Wang}}]{han2004modeling}%
  \BibitemOpen
  \bibfield  {author} {\bibinfo {author} {\bibfnamefont {M.}~\bibnamefont
  {Han}}, \bibinfo {author} {\bibfnamefont {Z.}~\bibnamefont {Shi}},\ and\
  \bibinfo {author} {\bibfnamefont {W.}~\bibnamefont {Wang}},\ }\bibfield
  {title} {\bibinfo {title} {Modeling dynamic system by recurrent neural
  network with state variables},\ }in\ \href@noop {} {\emph {\bibinfo
  {booktitle} {International Symposium on Neural Networks}}}\ (\bibinfo
  {organization} {Springer},\ \bibinfo {year} {2004})\ pp.\ \bibinfo {pages}
  {200--205}\BibitemShut {NoStop}%
\bibitem [{\citenamefont {Zimmermann}\ and\ \citenamefont
  {Neuneier}(2000)}]{zimmermann2000modeling}%
  \BibitemOpen
  \bibfield  {author} {\bibinfo {author} {\bibfnamefont {H.}~\bibnamefont
  {Zimmermann}}\ and\ \bibinfo {author} {\bibfnamefont {R.}~\bibnamefont
  {Neuneier}},\ }\bibfield  {title} {\bibinfo {title} {Modeling dynamical
  systems by recurrent neural networks},\ }\href@noop {} {\bibfield  {journal}
  {\bibinfo  {journal} {WIT Transactions on Information and Communication
  Technologies}\ }\textbf {\bibinfo {volume} {25}} (\bibinfo {year}
  {2000})}\BibitemShut {NoStop}%
\bibitem [{\citenamefont {Deyle}\ \emph {et~al.}(2013)\citenamefont {Deyle},
  \citenamefont {Fogarty}, \citenamefont {Hsieh}, \citenamefont {Kaufman},
  \citenamefont {MacCall}, \citenamefont {Munch}, \citenamefont {Perretti},
  \citenamefont {Ye},\ and\ \citenamefont {Sugihara}}]{Deyle2013}%
  \BibitemOpen
  \bibfield  {author} {\bibinfo {author} {\bibfnamefont {E.~R.}\ \bibnamefont
  {Deyle}}, \bibinfo {author} {\bibfnamefont {M.}~\bibnamefont {Fogarty}},
  \bibinfo {author} {\bibfnamefont {C.-h.}\ \bibnamefont {Hsieh}}, \bibinfo
  {author} {\bibfnamefont {L.}~\bibnamefont {Kaufman}}, \bibinfo {author}
  {\bibfnamefont {A.~D.}\ \bibnamefont {MacCall}}, \bibinfo {author}
  {\bibfnamefont {S.~B.}\ \bibnamefont {Munch}}, \bibinfo {author}
  {\bibfnamefont {C.~T.}\ \bibnamefont {Perretti}}, \bibinfo {author}
  {\bibfnamefont {H.}~\bibnamefont {Ye}},\ and\ \bibinfo {author}
  {\bibfnamefont {G.}~\bibnamefont {Sugihara}},\ }\bibfield  {title} {\bibinfo
  {title} {{Predicting climate effects on Pacific sardine}},\ }\href
  {https://doi.org/10.1073/pnas.1215506110} {\bibfield  {journal} {\bibinfo
  {journal} {Proceedings of the National Academy of Sciences}\ }\textbf
  {\bibinfo {volume} {110}},\ \bibinfo {pages} {6430} (\bibinfo {year}
  {2013})}\BibitemShut {NoStop}%
\bibitem [{\citenamefont {Munch}\ \emph {et~al.}(2018)\citenamefont {Munch},
  \citenamefont {Giron-Nava},\ and\ \citenamefont {Sugihara}}]{Munch2018}%
  \BibitemOpen
  \bibfield  {author} {\bibinfo {author} {\bibfnamefont {S.~B.}\ \bibnamefont
  {Munch}}, \bibinfo {author} {\bibfnamefont {A.}~\bibnamefont {Giron-Nava}},\
  and\ \bibinfo {author} {\bibfnamefont {G.}~\bibnamefont {Sugihara}},\
  }\bibfield  {title} {\bibinfo {title} {{Nonlinear dynamics and noise in
  fisheries recruitment: A global meta-analysis}},\ }\href
  {https://doi.org/10.1111/faf.12304} {\bibfield  {journal} {\bibinfo
  {journal} {Fish and Fisheries}\ }\textbf {\bibinfo {volume} {19}},\ \bibinfo
  {pages} {964} (\bibinfo {year} {2018})}\BibitemShut {NoStop}%
\bibitem [{\citenamefont {Morales}\ \emph {et~al.}(2004)\citenamefont
  {Morales}, \citenamefont {Haydon}, \citenamefont {Frair}, \citenamefont
  {Holsinger},\ and\ \citenamefont {Fryxell}}]{Morales2004a}%
  \BibitemOpen
  \bibfield  {author} {\bibinfo {author} {\bibfnamefont {J.~M.}\ \bibnamefont
  {Morales}}, \bibinfo {author} {\bibfnamefont {D.~T.}\ \bibnamefont {Haydon}},
  \bibinfo {author} {\bibfnamefont {J.}~\bibnamefont {Frair}}, \bibinfo
  {author} {\bibfnamefont {K.~E.}\ \bibnamefont {Holsinger}},\ and\ \bibinfo
  {author} {\bibfnamefont {J.~M.}\ \bibnamefont {Fryxell}},\ }\bibfield
  {title} {\bibinfo {title} {{Extracting more out of relocation data: Building
  movement models as mixtures of random walks}},\ }\bibfield  {journal}
  {\bibinfo  {journal} {Ecology}\ }\href {https://doi.org/10.1890/03-0269}
  {10.1890/03-0269} (\bibinfo {year} {2004})\BibitemShut {NoStop}%
\bibitem [{\citenamefont {Hinton}\ \emph {et~al.}(2012)\citenamefont {Hinton},
  \citenamefont {Srivastava},\ and\ \citenamefont
  {Swersky}}]{hinton2012neural}%
  \BibitemOpen
  \bibfield  {author} {\bibinfo {author} {\bibfnamefont {G.}~\bibnamefont
  {Hinton}}, \bibinfo {author} {\bibfnamefont {N.}~\bibnamefont {Srivastava}},\
  and\ \bibinfo {author} {\bibfnamefont {K.}~\bibnamefont {Swersky}},\
  }\bibfield  {title} {\bibinfo {title} {Neural networks for machine learning
  lecture 6a overview of mini-batch gradient descent},\ }\href@noop {}
  {\bibfield  {journal} {\bibinfo  {journal} {Unpublished lecture}\ }\textbf
  {\bibinfo {volume} {14}} (\bibinfo {year} {2012})}\BibitemShut {NoStop}%
\bibitem [{\citenamefont {Prechelt}(1998)}]{prechelt1998automatic}%
  \BibitemOpen
  \bibfield  {author} {\bibinfo {author} {\bibfnamefont {L.}~\bibnamefont
  {Prechelt}},\ }\bibfield  {title} {\bibinfo {title} {Automatic early stopping
  using cross validation: quantifying the criteria},\ }\href@noop {} {\bibfield
   {journal} {\bibinfo  {journal} {Neural Networks}\ }\textbf {\bibinfo
  {volume} {11}},\ \bibinfo {pages} {761} (\bibinfo {year} {1998})}\BibitemShut
  {NoStop}%
\bibitem [{\citenamefont {Lorenz}(1963)}]{lorenz1963deterministic}%
  \BibitemOpen
  \bibfield  {author} {\bibinfo {author} {\bibfnamefont {E.~N.}\ \bibnamefont
  {Lorenz}},\ }\bibfield  {title} {\bibinfo {title} {Deterministic nonperiodic
  flow},\ }\href@noop {} {\bibfield  {journal} {\bibinfo  {journal} {Journal of
  atmospheric sciences}\ }\textbf {\bibinfo {volume} {20}},\ \bibinfo {pages}
  {130} (\bibinfo {year} {1963})}\BibitemShut {NoStop}%
\bibitem [{\citenamefont {Duffing}(1918)}]{duffing1918erzwungene}%
  \BibitemOpen
  \bibfield  {author} {\bibinfo {author} {\bibfnamefont {G.}~\bibnamefont
  {Duffing}},\ }\href@noop {} {\emph {\bibinfo {title} {Erzwungene Schwingungen
  bei ver{\"a}nderlicher Eigenfrequenz und ihre technische Bedeutung}}},\
  \bibinfo {number} {41-42}\ (\bibinfo  {publisher} {Vieweg},\ \bibinfo {year}
  {1918})\BibitemShut {NoStop}%
\bibitem [{\citenamefont {Lorenz}(1996)}]{lorenz1996predictability}%
  \BibitemOpen
  \bibfield  {author} {\bibinfo {author} {\bibfnamefont {E.~N.}\ \bibnamefont
  {Lorenz}},\ }\bibfield  {title} {\bibinfo {title} {Predictability: A problem
  partly solved},\ }in\ \href@noop {} {\emph {\bibinfo {booktitle} {Proc.
  Seminar on predictability}}},\ Vol.~\bibinfo {volume} {1}\ (\bibinfo {year}
  {1996})\BibitemShut {NoStop}%
\bibitem [{\citenamefont {Lotka}(1910)}]{lotka1910contribution}%
  \BibitemOpen
  \bibfield  {author} {\bibinfo {author} {\bibfnamefont {A.~J.}\ \bibnamefont
  {Lotka}},\ }\bibfield  {title} {\bibinfo {title} {Contribution to the theory
  of periodic reactions},\ }\href {https://doi.org/10.1021/j150111a004}
  {\bibfield  {journal} {\bibinfo  {journal} {The Journal of Physical
  Chemistry}\ }\textbf {\bibinfo {volume} {14}},\ \bibinfo {pages} {271}
  (\bibinfo {year} {1910})},\ \Eprint
  {https://arxiv.org/abs/https://doi.org/10.1021/j150111a004}
  {https://doi.org/10.1021/j150111a004} \BibitemShut {NoStop}%
\bibitem [{\citenamefont {Chattopadhyay}\ \emph {et~al.}(2020)\citenamefont
  {Chattopadhyay}, \citenamefont {Hassanzadeh},\ and\ \citenamefont
  {Subramanian}}]{chattopadhyay2020datadriven}%
  \BibitemOpen
  \bibfield  {author} {\bibinfo {author} {\bibfnamefont {A.}~\bibnamefont
  {Chattopadhyay}}, \bibinfo {author} {\bibfnamefont {P.}~\bibnamefont
  {Hassanzadeh}},\ and\ \bibinfo {author} {\bibfnamefont {D.}~\bibnamefont
  {Subramanian}},\ }\bibfield  {title} {\bibinfo {title} {Data-driven
  predictions of a multiscale lorenz 96 chaotic system using machine-learning
  methods: reservoir computing, artificial neural network, and long short-term
  memory network},\ }\href {https://doi.org/10.5194/npg-27-373-2020} {\bibfield
   {journal} {\bibinfo  {journal} {Nonlinear Processes in Geophysics}\ }\textbf
  {\bibinfo {volume} {27}},\ \bibinfo {pages} {373} (\bibinfo {year}
  {2020})}\BibitemShut {NoStop}%
\bibitem [{\citenamefont {Dueben}\ and\ \citenamefont
  {Bauer}(2018)}]{dueben2018challenges}%
  \BibitemOpen
  \bibfield  {author} {\bibinfo {author} {\bibfnamefont {P.~D.}\ \bibnamefont
  {Dueben}}\ and\ \bibinfo {author} {\bibfnamefont {P.}~\bibnamefont {Bauer}},\
  }\bibfield  {title} {\bibinfo {title} {Challenges and design choices for
  global weather and climate models based on machine learning},\ }\href
  {https://doi.org/10.5194/gmd-11-3999-2018} {\bibfield  {journal} {\bibinfo
  {journal} {Geoscientific Model Development}\ }\textbf {\bibinfo {volume}
  {11}},\ \bibinfo {pages} {3999} (\bibinfo {year} {2018})}\BibitemShut
  {NoStop}%
\end{thebibliography}%

\newpage
\appendix
\begin{figure*}
    \centering
    \includegraphics[width=1.\textwidth]{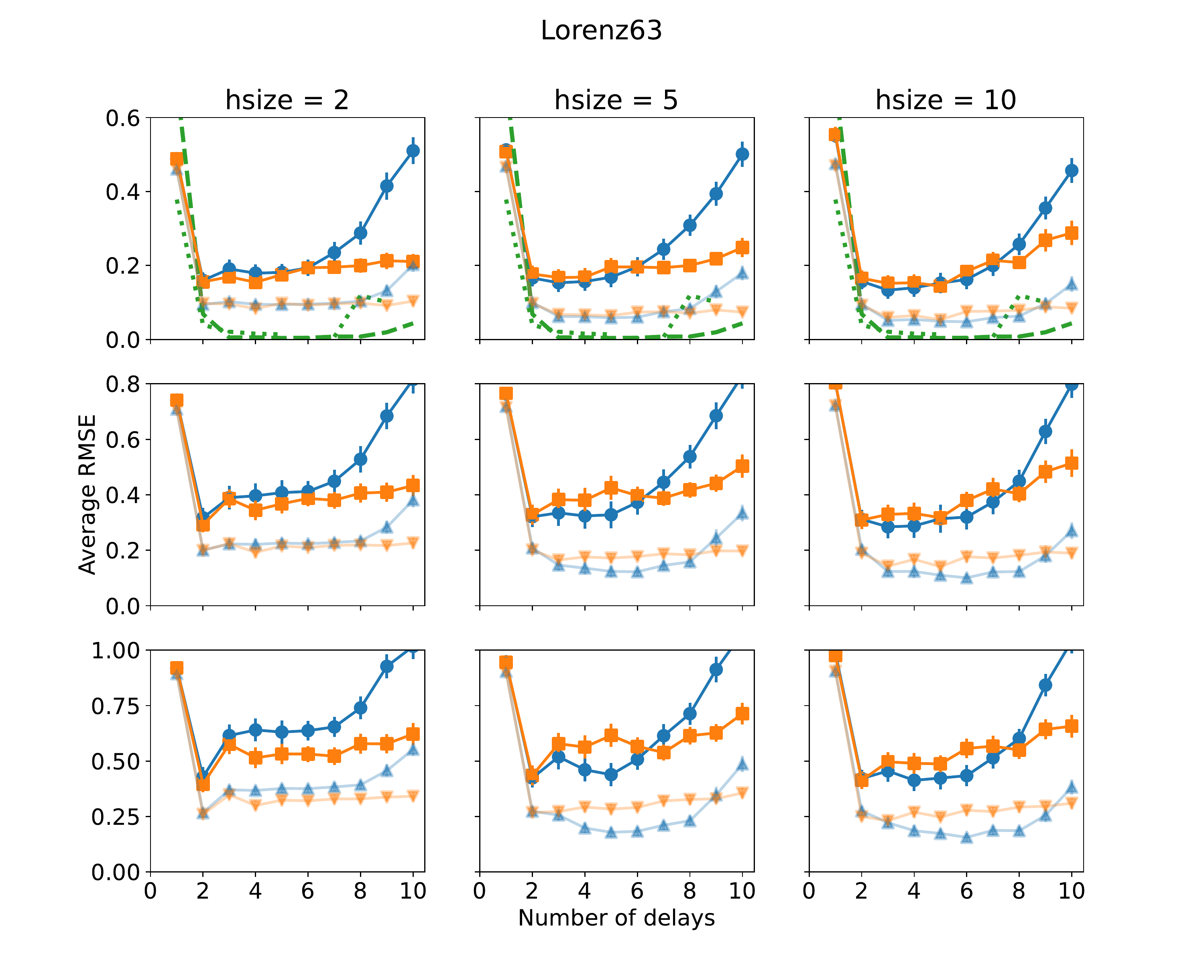}
    \caption{Normalized RMSE as a function of number of delays from a FNN (blue) vs RNN (orange) for the Lorenz63 model \eqref{lorenz63}. The top, middle, and bottom panels correspond to one-, two-, and three-steps ahead forecast error from a model trained on one-step ahead data. The left, middle, and right panels correspond to neural networks with hidden-layers with two, five and ten neurons each. The green dashed (dotted) lines in the top panel are numerical evaluations of the one-step ahead recursion error \eqref{apxd1} (and its first-order approximation \eqref{vard}). Dark (light) colors are results for with a training size of 50 (100) data points}
    \label{fig:results_l63}
\end{figure*}

\begin{figure*}
    \centering
    \includegraphics[width=1.\textwidth]{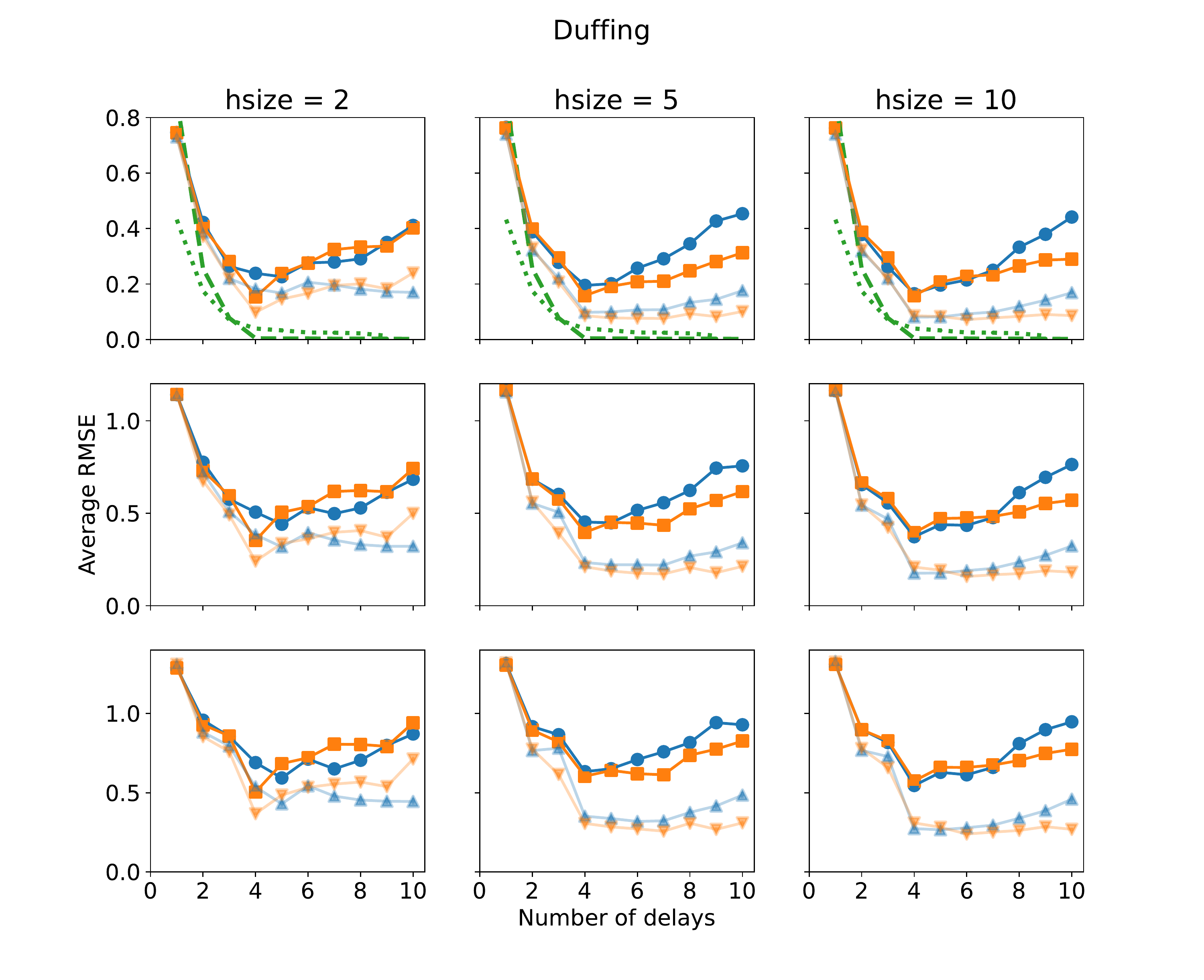}
    \caption{Normalized RMSE as a function of number of delays from a FNN (blue) vs RNN (orange) for the Lorenz63 model \eqref{duffing}. The top, middle, and bottom panels correspond to one-, two-, and three-steps ahead forecast error from a model trained on one-step ahead data. The left, middle, and right panels correspond to neural networks with hidden-layers with two, five and ten neurons each. The green dashed (dotted) lines in the top panel are numerical evaluations of the one-step ahead recursion error \eqref{apxd1} (and its first-order approximation \eqref{vard}). Dark (light) colors are results for with a training size of 50 (100) data points}
    \label{fig:results_df}
\end{figure*}

\begin{figure*}
    \centering
    \includegraphics[width=1.\textwidth]{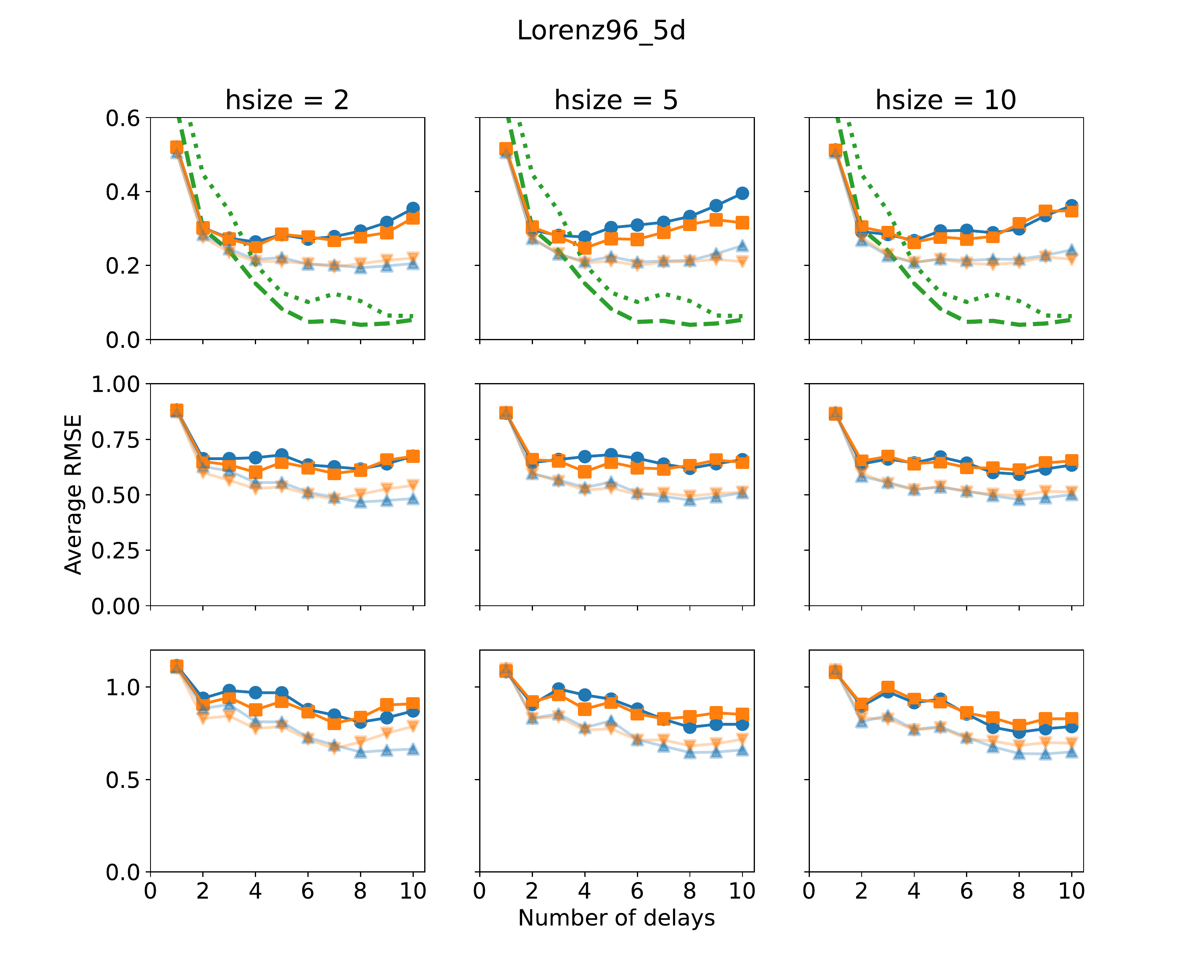}
    \caption{Normalized RMSE as a function of number of delays from a FNN (blue) vs RNN (orange) for the Lorenz63 model \eqref{lorenz96}. The top, middle, and bottom panels correspond to one-, two-, and three-steps ahead forecast error from a model trained on one-step ahead data. The left, middle, and right panels correspond to neural networks with hidden-layers with two, five and ten neurons each. The green dashed (dotted) lines in the top panel are numerical evaluations of the one-step ahead recursion error \eqref{apxd1} (and its first-order approximation \eqref{vard}). Dark (light) colors are results for with a training size of 50 (100) data points}
    \label{fig:results_l96}
\end{figure*}

\end{document}